\documentclass[pdflatex,sn-mathphys-num]{sn-jnl}%
\usepackage{capt-of}
\usepackage{caption}
\usepackage{chngcntr}

\makeatletter
\renewcommand{\thefigure}{%
  \ifnum\value{subsection}>0
    \thesection.\arabic{subsection}
  \else
    \thesection
  \fi}
\makeatother
\usepackage{xcolor}
\usepackage{soul}
\usepackage[table,xcdraw]{xcolor}
\usepackage{xspace}
\usepackage{CJKutf8}
\usepackage[T1]{fontenc}
\usepackage{svg}
\usepackage[most]{tcolorbox}
\usepackage{pifont}
\definecolor{softgreen}{RGB}{34, 149, 34}
\definecolor{softred}{RGB}{232, 34, 34}
\definecolor{correctgreen}{rgb}{0.0, 0.5, 0.0} 
\definecolor{wrongred}{rgb}{0.8, 0.0, 0.0} 
\newtheorem{takeaway}{Take-away}
\newcommand{\cmark}{\textcolor{softgreen}{\ding{51}}} 
\newcommand{\xmark}{\textcolor{softred}{\ding{55}}} 
\usepackage{threeparttable}
\usepackage{graphicx}
\usepackage{url}
\usepackage{booktabs}
\usepackage{url}
\usepackage{xurl}
\usepackage{graphicx}%
\usepackage{multirow}%
\usepackage{amsmath,amssymb,amsfonts}%
\usepackage{amsthm}%
\usepackage{mathrsfs}%
\usepackage[title]{appendix}%
\usepackage{xcolor}%
\usepackage{textcomp}%
\usepackage{manyfoot}%
\usepackage{booktabs}%
\usepackage{algorithm}%
\usepackage{algorithmicx}%
\usepackage{algpseudocode}%
\usepackage{listings}%
\usepackage[utf8]{inputenc}
\usepackage{textcomp}
\usepackage{caption}

\usepackage{xcolor}
\usepackage{soul}
\soulregister\cite7
\soulregister\ref7
\soulregister\eqref7

\definecolor{lightblue}{RGB}{200,220,255}

\definecolor{appendixbg}{RGB}{200,220,255}

\tcbset{
  aibox/.style={
    width=\textwidth,
    top=10pt,
    colback=white,
    colframe=black,
    colbacktitle=black,
    enhanced,
    center,
    attach boxed title to top left={yshift=-0.1in,xshift=0.15in},
    boxed title style={boxrule=0pt,colframe=white,},
  }
}
\newtcolorbox{AIbox}[2][]{aibox,title=#2,#1}

\theoremstyle{thmstyleone}%
\theoremstyle{thmstyletwo}%

\theoremstyle{thmstylethree}%

\begin{document}
\begin{CJK}{UTF8}{gbsn}
\title[Enabling Doctor-Centric Medical AI with LLMs through Workflow-Aligned Tasks and Benchmarks]{Enabling Doctor-Centric Medical AI with LLMs through Workflow-Aligned Tasks and Benchmarks}


\author[1,2]{\fnm{Wenya} \sur{Xie}}\email{xie00470@umn.edu}

\author[3]{\fnm{Qingying} \sur{Xiao}}\email{xiaoqingying@sribd.cn}

\author[1]{\fnm{Yu} \sur{Zheng}}\email{yuzheng@link.cuhk.edu.cn}

\author[1]{\fnm{Xidong} \sur{Wang}}\email{xidongwang1@link.cuhk.edu.cn}

\author[1,2]{\fnm{Junying} \sur{Chen}}\email{junyingchen2@link.cuhk.edu.cn}

\author[1]{\fnm{Ke} \sur{Ji}}\email{keji@link.cuhk.edu.cn}

\author[2]{\fnm{Anningzhe} \sur{Gao}}\email{anningzhegao@gmail.com}

\author[4]{\fnm{Prayag} \sur{Tiwari}}\email{prayag.tiwari@ieee.org}

\author[2]{\fnm{Xiang} \sur{Wan}}\email{wanxiang@sribd.cn}

\author*[1,2,5]{\fnm{Feng} \sur{Jiang}}\email{jiangfeng@suat-sz.edu.cn}

\author*[1,2]{\fnm{Benyou} \sur{Wang}}\email{wangbenyou@cuhk.edu.cn}

\affil[1]{%
  \orgdiv{School of Data Science},%
  \orgname{The Chinese University of Hong Kong, Shenzhen},%
  \orgaddress{%
    \street{2001 Longxiang Boulevard, Longgang District},%
    \city{Shenzhen},%
    \postcode{518172},%
    \state{Guangdong},%
    \country{China}}}

\affil[2]{%
  \orgname{Shenzhen Research Institute of Big Data},%
  \orgaddress{%
    \street{2001 Longxiang Boulevard, Longgang District},%
    \city{Shenzhen},%
    \postcode{518172},%
    \state{Guangdong},%
    \country{China}}}

\affil[3]{%
  \orgname{National Health Data Institute, Shenzhen},%
  \orgaddress{%
    \street{2001 Longxiang Boulevard, Longgang District},%
    \city{Shenzhen},%
    \postcode{518172},%
    \state{Guangdong},%
    \country{China}}}

\affil[4]{%
  \orgname{Halmstad University},%
  \orgaddress{%
    \street{PO Box 823, Kristian IV:s väg 3},%
    \city{Halmstad},%
    \postcode{301 18},%
    \state{Halland},%
    \country{Sweden}}}

\affil[5]{%
  \orgname{Shenzhen University of Advanced Technology},%
  \orgaddress{%
    \street{No.1 Gongchang Road, Guangming District, },%
    \city{Shenzhen},%
    \postcode{518107},%
    \state{Guangdong},%
    \country{China}}}
\abstract{
The rise of large language models (LLMs) has transformed healthcare by offering clinical guidance, yet their direct deployment to patients poses safety risks due to limited domain expertise. To mitigate this, we propose repositioning LLMs as clinical assistants that collaborate with experienced physicians rather than interacting with patients directly. We conduct a two-stage inspiration–feedback survey to identify real-world needs in clinical workflows. Guided by this, we construct DoctorFLAN, a large-scale Chinese medical dataset comprising 92,000 Q\&A instances across 22 clinical tasks and 27 specialties. To evaluate model performance in doctor-facing applications, we introduce DoctorFLAN-test (550 single-turn Q\&A items) and DotaBench (74 multi-turn conversations). Experimental results with over ten popular LLMs demonstrate that DoctorFLAN notably improves the performance of open-source LLMs in medical contexts, facilitating their alignment with physician workflows and complementing existing patient-oriented models. This work contributes a valuable resource and framework for advancing doctor-centered medical LLM development.}





\maketitle
\section{Introduction}\label{sec1}
Large Language Models (LLMs) have demonstrated significant potential in various applications within healthcare, such as autonomous online consultations, which can reduce costs and improve accessibility to medical services \cite{singhal2023large, singhal2025toward, li2023chatdoctor, wang2023clinicalgpt, chenhuatuogpt, zhang2023huatuogpt, wang2023huatuo, chen2023bianque}. 
However,  using LLMs as a direct consulting tool for patients can bring serious health risks because patients lacking medical expertise are easily misled by the inaccurate medical advice generated by the model~\cite{pal2023med, wang-etal-2024-cmb, fan2023quantifying}.

\begin{figure}
    \centering
    \includegraphics[width=0.8\textwidth]{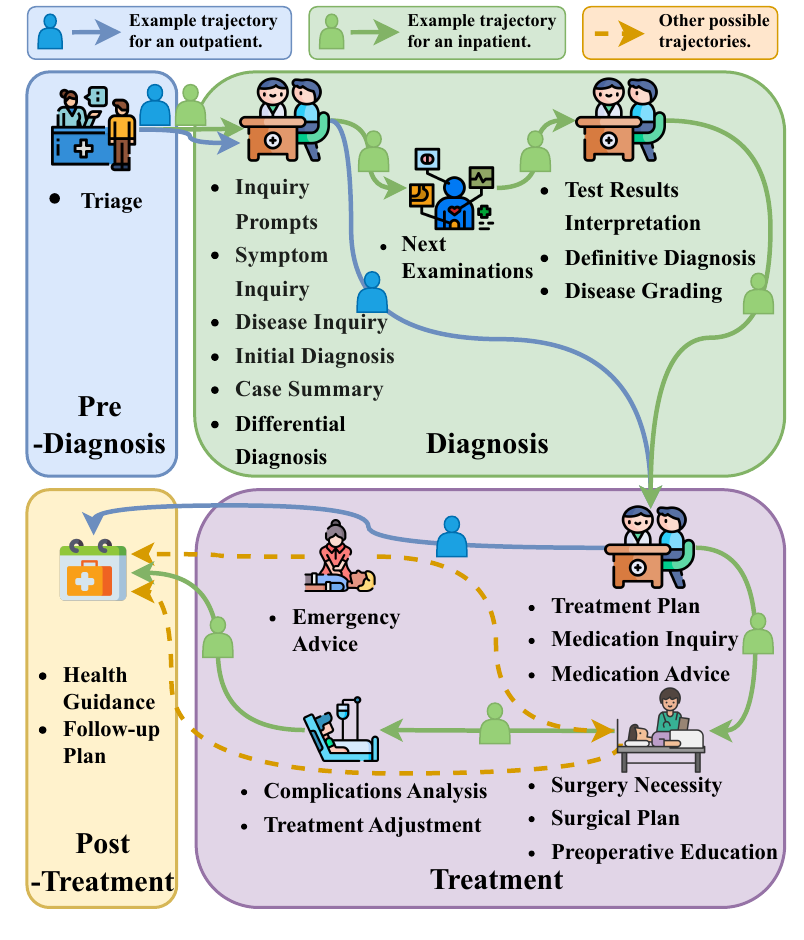}
    \refstepcounter{figure}\label{fig:full_process}
\end{figure}
In contrast, developing LLMs as medical assistants for healthcare professionals presents a safer and more practical direction. Doctors routinely deal with complex information processing tasks, such as summarizing patient records, providing clinical decision support, and educating patients. Using LLMs for these tasks could significantly alleviate the workload of doctors, allowing them to perform their duties more efficiently \cite{wu2023medical, chen2024dia}. Furthermore, large language models have shown promising results in multi-task settings\cite{wei2022finetuned, wang-etal-2022-super}, suggesting that LLMs have substantial potential when applied to a multi-functional medical assistant role.
Despite these promising developments, there remains a significant gap between the current capabilities of LLMs and the complex requirements of real-world medical practice. Most existing medical LLMs \cite{zhang2023huatuogpt, li2023chatdoctor, zeng2020meddialog, he2019applying, zhang2022cblue} have been trained on patient-centric datasets, which focus primarily on tasks like pre-diagnosis and medical consultation. These datasets are limited in scope and do not encompass the diverse and multifaceted nature of clinical tasks encountered in actual medical environments. Moreover, previous research on LLMs as medical assistants has often focused on a narrow set of tasks \cite{wu2023medical, chen2024dia}, and these models frequently fail to provide comprehensive responses to complex, real-world medical inquiries \cite{zhang2018multi, pal2022medmcqa}.
Another critical limitation lies in the current benchmark tests, which often do not adequately assess the performance of LLMs as medical assistants. Most widely used benchmarks rely on multiple-choice question formats \cite{jin2019pubmedqa, jin2021disease, singhal2023large, pal2022medmcqa, zhang2018multi, wang-etal-2024-cmb}, which fail to align with the real-world requirements where detailed and comprehensive responses are needed. Alternatively, these benchmarks typically assess only a small subset of tasks \cite{singhal2023large}, failing to cover the full range of workflows that doctors encounter in practice.

To address the above issues, we aim to develop LLMs as better doctor assistants by building comprehensive and practical datasets and evaluations. Firstly, to gain a thorough understanding of doctors’ needs for medical assistants, we collaborate with dozens of professional doctors to explore 22 tasks across four phases in real-world scenarios. These tasks are finalized through a two-stage survey using a heuristic-feedback method, as shown in Figure~\ref{fig:full_process}. Based on these insights, we develop DoctorFLAN, a comprehensive Chinese medical dataset containing approximately 92K samples that capture the full spectrum of the doctor's daily work, including both inpatient and outpatient scenarios. It leverages GPT-4-polishing with reference enhancement, followed by manual verification from professional doctors, to ensure samples provide reliable and comprehensive expert responses for training our model (DotaGPT).

To develop effective doctor assistants, we construct a novel benchmark for medical LLMs that includes both single-turn evaluation (DoctorFLAN-\textit{test}) and multi-turn evaluation (DotaBench) by simulating the dialogue with doctors in receiving patients scenarios. The existing popular LLMs and our model, DotaGPT, are evaluated both automatically and manually on the benchmark. The results indicate that existing models, while acting as virtual doctors assisting patients, struggle with the diverse and complex tasks required for real-world roles that assist doctors. In contrast, our DoatGPT serving as a doctor assistant exhibits robust performance across tasks in both DoctorFLAN-\textit{test} and DotaBench. 

Our contributions are threefold: 

(1) We explore the underexplored scenario of developing medical models as doctor assistants, providing essential data, models, and benchmarks that complement existing research in this domain.

(2) We construct a 92K-sample dataset for doctor assistants by collaborating with dozens of medical professionals, using a heuristic feedback method to identify 22 key tasks and employing reference-enhanced polishing and manual verification.

(3) We introduce an expert-involved benchmark to assess large language models in doctor-oriented scenarios, covering both single-turn and multi-turn interactions, and thoroughly analyze the consistency between manual and automatic evaluations, comparing them with widely accepted benchmarks.
\section{Result}
\subsection{Automatic Evaluation Results}

\begin{table}[t]
\centering
\small
\resizebox{\textwidth}{!}{%
\setlength{\tabcolsep}{2pt}
\begin{tabular}{lc|ccccc|c}
\toprule
\multirow{3}{*}{\textbf{Model}} & \multirow{3}{*}{\textbf{Size}} & \multicolumn{5}{c|}{\textbf{DoctorFLAN-\textit{test}}} & \textbf{DotaBench} \\
\cmidrule(lr){3-6}
 & & \textbf{Pre} & \multirow{2}{*}{\textbf{Diagnosis}} & \multirow{2}{*}{\textbf{Treatment}} &  \textbf{Post} &  \multirow{2}{*}{\textbf{Average}} & \multirow{2}{*}{\textbf{Average}} \\
  & & \textbf{-Diagnosis} &   &  &  \textbf{-Treatment} &   &  \\
\midrule
\rowcolor{gray!10}  \multicolumn{8}{c}{Open-source General LLMs} \\ \midrule
Qwen-1.8B-Chat & 1.8B & 5.28 & 4.56 & 3.96 & 5.44 & 4.48 & 4.68\\
Baichuan-13B-Chat & 13B & 6.20 & 6.51 & 6.31 & 7.55 & 6.57 & 7.59 \\
Baichuan2-7B-Chat & 7B & 6.32 & 6.36 & 6.34 & 7.70 & 6.59 & 7.41 \\
Baichuan2-13B-Chat & 13B & 6.76 & 6.85 & 6.94 & 7.81 & 7.04 & 7.47 \\
Yi-6B-Chat & 6B & 7.00 & 6.83 & 6.83 & 7.66 & 6.98 & 8.25\\
Yi-34B-Chat& 34B  & 7.36 & 7.38 & 7.95 & 8.78 & 7.80 & 8.65\\
\midrule
\rowcolor{gray!10}  \multicolumn{8}{c}{Open-source Medical LLMs} \\ \midrule
BianQue-2 & 6B & 5.56 & 3.27 & 3.65 & 4.78 & 3.72 & 4.12 \\
DISC-MedLLM& 13B & 5.56 & 4.23 & 3.54 & 5.14 & 4.24 & 4.97 \\
HuatuoGPT & 7B  & 5.32 & 4.24 & 3.72 & 4.92 & 4.29 & 5.88\\
HuatuoGPT-II& 7B  & 7.60 & 7.02 & 6.69 & 7.42 & 7.03 & 7.90\\
\textbf{DotaGPT}\textsubscript{Yi-6B}& 6B & 8.32 & ~~~~~~~~7.62\textsubscript{\color{wrongred} $\uparrow$ 11.6\%} & ~~~~~~~~7.68\textsubscript{\color{wrongred} $\uparrow$ 12.4\%} & 8.44 & ~~~~~~~~7.81\textsubscript{{\color{wrongred} $\uparrow$ 11.9\%}} & ~~~~~~~~8.36 \textsubscript{{\color{wrongred} $\uparrow$ 1.3\%}}\\
\textbf{DotaGPT}\textsubscript{Baichuan2-7B}& 7B & 8.48 & ~~~~~~~~8.01\textsubscript{\color{wrongred} $\uparrow$ 25.9\%} & ~~~~~~~~8.23\textsubscript{\color{wrongred} $\uparrow$ 29.8\%} & 8.80 & ~~~~~~~~8.25\textsubscript{{\color{wrongred} $\uparrow$ 25.2\%}} &
~~~~~~~~~8.36 \textsubscript{{\color{wrongred} $\uparrow$ 12.8\%}}\\ \midrule
\rowcolor{gray!10} \multicolumn{8}{c}{Proprietary LLMs} \\ \midrule
GPT-3.5 & N/A & 6.40 & 6.85 & 6.26 & 6.74 & 6.64 & 7.83\\
Claude-3 & N/A & 7.80 & 8.38 & 8.28 & 8.76 & 8.38 & 9.21\\
GPT-4 & N/A & 8.00 & 8.41 & 8.28 & 9.04 & 8.42 & 9.41\\ 
\bottomrule
\end{tabular}
}
\caption{Automatic Evaluation Results on DoctorFLAN-\textit{test} and DotaBench. The subscript of \textbf{DotaGPT} (e.g., \textbf{DotaGPT}\textsubscript{Yi-6B}) indicates the backbone on which the model was initially trained. The red arrows ({\color{wrongred} $\uparrow$}) with percentages indicate the improvement of DotaGPT over the corresponding chat models with the same backbone.}

\label{tab:auto eval-DoctorFLAN-DotaBench}
\end{table}
Table \ref{tab:auto eval-DoctorFLAN-DotaBench} outlines the automatic evaluation results of the existing medical models on DoctorFLAN-\textit{test}.
\begin{takeaway}
Existing models perform poorly in the Diagnosis and Treatment Phases.
\end{takeaway}
The results reveal a notable performance decline for all models during the diagnosis and treatment phases compared to the pre-diagnosis and post-treatment phases. This drop may be attributed to the high medical knowledge requirements of tasks like \textit{Disease Grading} and \textit{Surgical Plan}, for which models are often undertrained due to a lack of knowledge-intensive datasets. However, DotaGPT models show a significant improvement in these phases. Specifically, DotaGPT\textsubscript{Baichuan2-7B} and DotaGPT \textsubscript{Yi-6B} exhibit performance increases of 11.6\% and 12.4\% in the diagnosis phase, and 25.9\% and 29.8\% in the treatment phase, respectively. These enhancements demonstrate the value of our tailored dataset in improving performance on complex medical tasks.

\begin{takeaway}
Larger Models Perform Better.
\end{takeaway}
When comparing Yi-6B-Chat (average score: 6.98) with Yi-34B-Chat (average score: 7.80), and Baichuan2-7B-Chat (average score: 6.59) with Baichuan2-13B-Chat (average score: 7.04), we observe that larger models consistently outperform their smaller counterparts across all four phases. The models with more parameters achieve higher average scores, likely due to their enhanced reasoning abilities, which better equip them to handle the considerable complexity of the tasks in our evaluation.

\begin{takeaway}
Limitations of Virtual Doctor Models in Workflow Assistance Tasks.
\end{takeaway}
Virtual doctor models originally designed to provide medical advice to patients, such as BianQue-2 and HuatuoGPT, perform relatively poorly in tasks related to doctor workflow assistance, with scores of 4.12 and 5.88, respectively. These models are primarily trained on large medical dialogue datasets, where the focus is on mimicking the question-and-answer style of doctors, with the goal of functioning as a virtual doctor. However, medical dialogues like Huatuo26M\cite{wang2025huatuo} are mostly based on online consultations, which may not capture the full range of tasks involved in a doctor's workflow. As a result, these models struggle with more specific, nuanced tasks that occur in everyday medical practice.

\begin{takeaway}
Medical Dataset Fine-Tuning Does Not Always Enhance Performance on DoctorFLAN.
\end{takeaway}
A comparison between the DISC-MedLLM (4.24) and its chat counterpart, Baichuan-13B-Chat (6.57), reveals that the medical domain-specific fine-tuning of DISC-MedLLM does not lead to better performance on the DoctorFLAN tasks. In fact, the fine-tuned DISC-MedLLM underperforms compared to the general-purpose Baichuan-13B-Chat. This outcome underscores the potential risks of excessive specialization, suggesting that a balance between domain-specific fine-tuning and general adaptability is crucial for ensuring broader model applicability.

\begin{takeaway}
DoctorFLAN Fine-Tuning Improves Performance on Doctor-Assistance Tasks.
\end{takeaway}
In contrast, our DotaGPT variants, fine-tuned on the DoctorFLAN dataset, demonstrate significant performance improvements over their respective chat model counterparts. Specifically, the variant fine-tuned on Baichuan2-7B shows a substantial improvement of 25.2\%. Similarly, the DotaGPT variant fine-tuned on Yi-6B outperforms the Yi-6B-Chat by 11.9\%. The improvement on both backbones highlights the effectiveness of DoctorFLAN and brings our models' performance close to those of leading proprietary models such as Claude-3 and GPT-4.

We further evaluate DotaGPT's performance on DotaBench to assess its ability in practical multi-turn settings, which reflects its real-world applicability. This out-of-domain evaluation is detailed in Table \ref{tab:auto eval-DoctorFLAN-DotaBench}. 
Notably, our DotaGPT variants significantly outperform models of comparable size on DotaBench, even surpassing the larger Yi-34B-Chat model. This strong performance underscores DotaGPT's robust ability to generalize from DoctorFLAN to out-of-domain contexts.

\subsection{Human Evaluation Results}
\begin{table}[t]
\centering
\resizebox{0.6\linewidth}{!}{
\begin{tabular}{lc}
\toprule
\textbf{Models} & \textbf{Average Score} \\
\midrule
BianQue-2 & 4.58 \\
HuatuoGPT & 4.97 \\
DISC-MedLLM & 5.36 \\
Baichuan2-7B-Chat & 6.69 \\
GPT-4 & 8.06 \\
\textbf{DotaGPT}\textsubscript{Baichuan2-7B} & 7.83 \\
\bottomrule
\end{tabular}
}
\vspace{2mm}
\caption{Human Evaluation Results on DoctorFLAN-\textit{test}. For detailed task-by-task results.}
\label{tab:average_scores}
\end{table}

\begin{table}[ht]
\centering
\resizebox{0.6\linewidth}{!}{
\begin{tabular}{lc}
\toprule
\textbf{Model}  & \textbf{Average Score}  \\ \midrule
Baichuan2-7B-Chat & 8.25 \\
\textbf{DotaGPT}\textsubscript{Baichuan2-7B}  & ~~~~~~~8.54\textsubscript{\color{wrongred} $\uparrow$ 3.5\%}\\ 
\bottomrule
\end{tabular}%
}
\vspace{2mm}
\caption{Human Evaluation Results on the DotaBench. }

\label{tab: expert eval}
\end{table}
\begin{figure}[htbp]
    \centering
    \includegraphics[width=0.9\columnwidth]{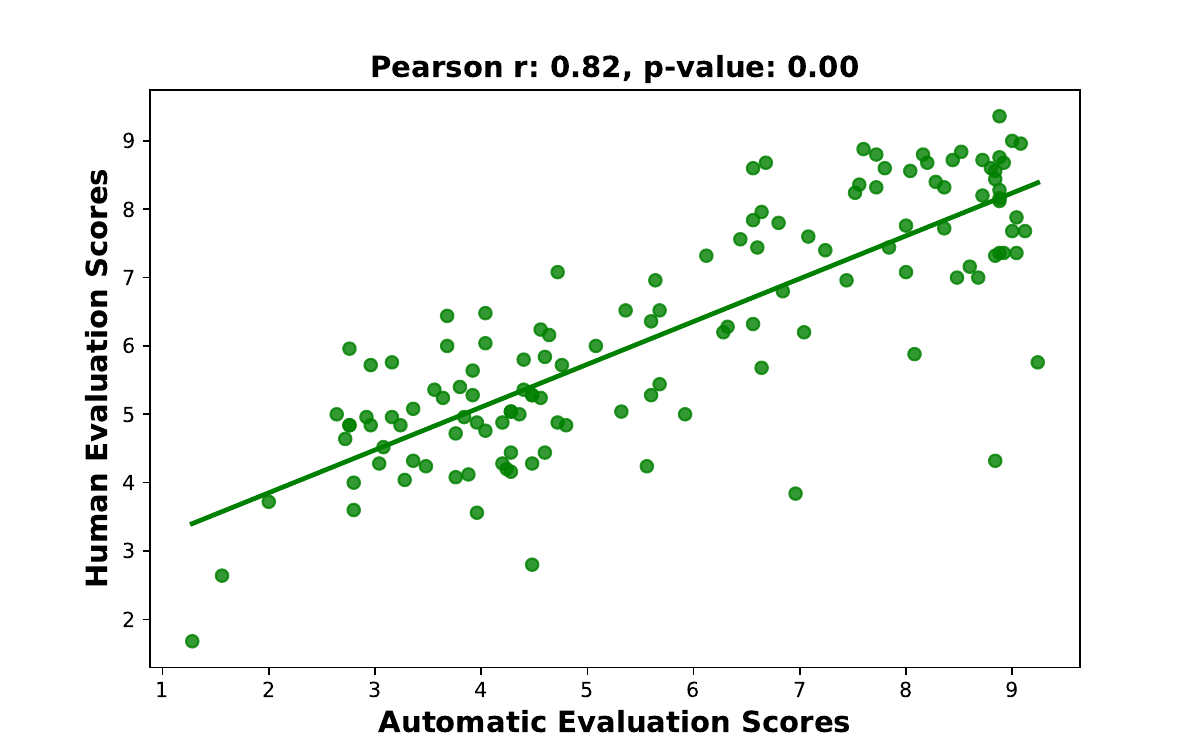}
    \refstepcounter{figure}\label{fig:correlation}
\end{figure}
Aside from the automatic evaluation, we conduct manual evaluation on a subset of models due to resource constraints, as shown in Tables 7 and 8. The results show that on DoctorFLAN-\textit{test}, DotaGPT\textsubscript{Baichuan2-7B} (7.83) outperforms patient-assistance models like BianQue-2 (4.58), HuatuoGPT (4.97), and DISC-MedLLM (5.36), as well as the general counterpart Baichuan2-7B-Chat (6.69), consistent with the automatic evaluation results.  
Further, human evaluation results on DotaBench, shown in Table \ref{tab: expert eval}, confirm DotaGPT\textsubscript{Baichuan2-7B}’s strong performance, with an average score of 8.54, surpassing Baichuan2-7B-Chat (8.25) by 3.5\%.

To verify the reliability of our evaluation methods, we also conduct a task-level correlation analysis between human and automatic evaluations on the DoctorFLAN-\textit{test}. For each model and task, we average the results across 25 samples per task (this averaging is done to ensure consistency and minimize the impact of outliers or variance in individual responses). Our analysis, covering 132 data points, reveals a Pearson correlation coefficient of \textbf{0.82}, indicating strong consistency between evaluation modes~\cite{zheng2024judging}, as shown in Figure \ref{fig:correlation}. 

\subsection{Generalization of DotaGPT on other benchmarks}
\begin{table}[ht]
\centering
\resizebox{0.8\linewidth}{!}{
\begin{tabular}{lcccc}
\toprule
\multirow{2}{*}{\textbf{Model}} & \multirow{2}{*}{\textbf{CMMLU}\textsubscript{\textit{Med.}}} & \multirow{2}{*}{\textbf{CMExam}} & \multirow{2}{*}{\textbf{MMLU}\textsubscript{\textit{Med.}}} & \textbf{CMB-} \\
&&&&\textbf{Exam}\\ \midrule
\rowcolor{gray!10} \multicolumn{5}{c}{Open-source Medical LLMs} \\ \midrule
DISC-MedLLM    & -              & 36.62           & -             & 32.47             \\
HuataoGPT-II   & 59.08          & 65.81           & 51.44         & 59.00             \\ 
Baichuan2-7B-Chat & 50.74       & 50.48           & 50.29         & 43.33             \\
\textbf{DotaGPT}\textsubscript{Baichuan2-7B} & 54.58 & 59.76 & 48.49 & ~~52.42$^*$      \\ \midrule
\rowcolor{gray!10} \multicolumn{5}{c}{Proprietary LLMs} \\ \midrule
GPT-4          & -              & -               & -             & 59.46             \\ 

\bottomrule
\end{tabular}
}
\vspace{2mm}
\caption{Comparative Performance of Medical LLMs on Diverse Medical Benchmarks. CMB-Exam scores are from \cite{wang-etal-2024-cmb}, except for DotaGPT\textsubscript{Baichuan2-7B}\textsuperscript{*}.}
\label{tab:model_comparison}
\end{table}
To further evaluate DotaGPT's medical knowledge and generalization capability, we assess its performance on several established medical benchmarks, as shown in Table \ref{tab:model_comparison}. DotaGPT\textsubscript{Baichuan2-7B} delivers competitive results across CMMLU~\cite{li-etal-2024-cmmlu}, MMLU~\cite{hendrycksmeasuring}, CMExam~\cite{NEURIPS2023_a48ad12d}, and CMB-Exam~\cite{wang-etal-2024-cmb}. Notably, it outperforms Baichuan2-7B-Chat in 3 out of 4 categories. Although DotaGPT\textsubscript{Baichuan2-7B} falls short of HuatuoGPT-II, this performance gap may be attributed to the significantly larger training dataset used by HuatuoGPT-II.

\subsection{Case Study}

\begin{table*}[h!]
\centering
\resizebox{\linewidth}{!}{
{\scriptsize 
\begin{tabular}{l}
\toprule
\textbf{Query:} \\
基于病人的病史、症状和检查结果，先列出几个可能的诊断，再解释为什么这些诊断可能与病人的病情相符，\\以及如何通过进一步的检查或观察来确认最终诊断。<病例描述>：患者，男，11个月，家长主诉发现患儿右眼中央白色反光2周，\\无眼红，无斜视。家族中无此类病者，否认外伤史，系足月顺产，无吸氧史，无产伤史。\\
(Translated)
Based on the patient's history, symptoms, and examination results, list several possible diagnoses first, 
then explain why these \\diagnoses may correspond to the patient's condition and how to confirm the final diagnosis through further examination or observation. \\
\textless Case description\textgreater: \\
The patient was a male, 11 months old, whose parents complained of a 2-week-old \textbf{white reflection in the centre of the right eye, }\\
\textbf{with no redness of the eye and no}\textbf{strabismus}. There was no family history of this condition, and the patient denied any history of \\trauma. He had a full-term delivery with no history of oxygen intake and no history of birth injury.\\\midrule
\textbf{Golden:}\\
基于患儿的病史及症状，以下是几个可能的初步诊断：\\
1. 先天性白内障 - 患儿年龄较小，无其他病史如早产等，白色反光可能是由于晶状体混浊造成。\\
2. 视网膜母细胞瘤（Retinoblastoma） - 虽然家族中无此类病例，虽然家族中无此类病例，但这是儿童最常见的原发性眼内恶性肿瘤，\\通常表现为眼底白色反光...
为何这些诊断可能符合病人病情：\\- 先天性白内障：眼底反光可能是由于光源照射到混浊的晶状体产生的，无其他临床症状如红眼或斜视，意味着可能是先天性问题。\\- 视网膜母细胞瘤：患儿的白色眼底反光（也称为猫眼反射）是视网膜母细胞瘤的典型表现。\\- 考虑PHPV等发育异常：通常发生在单侧，可以出现白色眼底反光，且不伴有红眼。\\- 视网膜炎也可能导致类似症状，但其他炎症的征兆通常会伴随出现...
\\
\textbf{(Translated)}
Based on the history and symptoms of the child, the following are a few possible initial diagnoses:\\1. Congenital cataract - The child is young and has no other medical history such as prematurity...\\2. Retinoblastoma (Retinoblastoma) - Although there is no such case in the family...\\
...\\
Why these diagnoses may be consistent with the patient's condition: \\- Congenital cataract: fundus reflections may be produced by a light source hitting the cloudy lens.. \\- Retinoblastoma: white fundus reflections (also known as cat's eye reflex) in children are typical of retinoblastoma. \\- Consider developmental anomalies such as PHPV: usually occurring unilaterally...\\\midrule
\textbf{DotaGPT\textsubscript{Baichuan2-7B}:}\\
根据您提供的病历信息，以下是一些可能的初步诊断及其合理性分析：\\ 可能诊断: 1. **先天性白内障** (Congenital Cataract):\\
- **理由**: 白内障通常表现为瞳孔区或晶状体位置的白色反光...
\\2. **视网膜母细胞瘤** (Retinoblastoma):
\\   - **理由**: 在婴幼儿中，视网膜母细胞瘤是眼部最常见的恶性肿瘤。尽管通常伴有眼红和斜视等症状，但早期病变也可能只表现为白色反光...\\
Based on the information you have provided in your medical history, here are some possible preliminary diagnoses and their plausibility:\\
Possible Diagnoses:\\
1. \color{correctgreen}{Congenital Cataract}\cmark:\\ - Rationale: \color{correctgreen}{A cataract usually presents as a white reflection in the pupil area or at the location of the lens...} \cmark  \\ - FURTHER EXAMINATION: Ultrasound may be performed to assess whether the structure and position of the lens is normal...\\ 
2. \color{correctgreen}{Retinoblastoma (Retinoblastoma)}\cmark:\\ - Rationale: In infants and children, retinoblastoma is the most common malignant tumour of the eye. \color{correctgreen}{Although usually associated with} \\\color{correctgreen}{symptoms such as redness of the eye and strabismus, early lesions may also present only as white reflections.}\cmark\\...\\\midrule
\textbf{HuatuoGPT:}\\
根据您提供的病史，以下是一些可能的诊断：\\1. 结膜炎：结膜炎是一种常见的眼部疾病，通常由病毒或细菌感染引起。患者的眼睛可能会出现充血、分泌物增多、疼痛等症状。如果您的孩子\\没有其他症状，如发热、咳嗽等，那么结膜炎是一个可能的诊断。\\2. 白内障：白内障是一种常见的老年眼病，通常会导致晶状体混浊，影响视力。如果您的孩子没有其他症状，如头痛、呕吐等，那么白内障\\也是一个可能的诊断。\\3. 角膜炎：角膜炎是一种炎症性疾病，通常由细菌感染引起。患者的眼睛可能会出现充血、疼痛、流泪等症状。如果您的孩子没有其他症状，\\如发热、咳嗽等，那么角膜炎也是一个可能的诊断。\\4. 青光眼：青光眼是一种常见的眼病，通常导致眼压升高，损伤视神经，影响视力。如果您的孩子没有其他症状，如头痛、呕吐等，那么青光眼\\也是一个可能的诊断...\\
\textbf{(Translated)}
Based on the history you have provided, here are some possible diagnoses:\\
1. Conjunctivitis: Conjunctivitis is a common eye condition usually caused by a viral or bacterial infection... \\
2. Cataracts: Cataracts are a common eye disease in the elderly, usually resulting in clouding of the lens and affecting vision. \\Cataracts are \color{wrongred}{also a possible diagnosis if your child has no other symptoms such as fever or cough}, such as headaches or vomiting.\xmark\\ 
3. Keratitis: Keratitis is an inflammatory disease ... Keratitis is \color{wrongred}{also a possible diagnosis if your child has no other symptoms such as}\\ \color{wrongred}{fever or cough...}\xmark \\ 
4. Glaucoma: Glaucoma is a common eye disease ... Glaucoma is \color{wrongred}{also a possible diagnosis if your child has no other symptoms such as}\\\color{wrongred}{fever or cough...} \xmark\\\bottomrule
\end{tabular}
}
}
\caption{Illustrative case study from the \textit{Initial Diagnosis} task in DoctorFLAN-\textit{test}, showing Chinese model responses along with key English highlights for clarity. The example includes the ground truth (Golden) and outputs from DotaGPT\textsubscript{Baichuan2-7B} and HuatuoGPT. We annotate key segments using \textcolor{correctgreen}{green} \cmark for medically correct information and \textcolor{wrongred}{red} \xmark for incorrect or irrelevant reasoning. The full Chinese outputs are preserved to support fine-grained comparison across models.}
\label{tab:case_study}
\end{table*}
To provide a clearer demonstration of our model's ability to generate knowledge-intensive responses in doctor-oriented tasks, we select a case from \textit{Differential Diagnosis} for comparison. As detailed in Table \ref{tab:case_study}, we present the responses from both HuatuoGPT and DotaGPT(Baichuan2-7B).
Despite HuatuoGPT's fluent responses, they lack substantial information, often repeating general rather than medically specific answers. Conversely, DotaGPT not only accurately diagnosed the condition but also provided detailed, professional reasoning, demonstrating its superior ability to deliver knowledge-intensive answers.
\section{Discussion}

In this paper, we focus on underexplored scenarios of developing medical models as doctor assistants. We first collaborate with dozens of doctors and conduct a two-stage survey to accurately identify real-world clinical tasks for efficient doctor assistance. We then create DoctorFLAN, using reference-enhanced refinement to overcome the training limitations of previous models. Additionally, we introduce DotaBench as a complementary evaluation to assess the effectiveness of popular medical LLMs as doctor assistants. Benchmark results indicate that while existing LLMs face challenges in this role, DotaGPT's performance shows that our dataset can significantly enhance their capability, providing a valuable supplement to current medical LLM research.

We also acknowledge the following limitations of this stage work. The DoctorFLAN is currently only available in Chinese and may require supplementation in other languages. Consequently, it cannot be guaranteed that DotaGPT trained on DoctorFLAN will perform well in languages other than the one on which it has been tested. However, the methodology employed to create DoctorFLAN can be applied universally across different languages. Additionally, although DotaGPT has demonstrated impressive performance on the benchmarks, it is important to exercise caution when using its outputs, particularly in real-world doctor-oriented interactions.

Ensuring the privacy and security of data is paramount in the development of medical applications. The datasets used in this study, primarily derived from Medtiku, an open-source repository of medical examination questions, and PromptCBLUE~\cite{zhu2023promptcblue}, are both freely available for use. Additionally, we incorporate data from an internet medical encyclopedia hosted by 120 Ask~\cite{ask120}, also open to the public.

Given the potential issues with the credibility of content generated by DotaGPT, we are committed to strictly regulating the model's use to prevent misuse. Our datasets, DoctorFLAN and DotaBench, will be released under terms that uphold the highest ethical standards. This commitment ensures that while advancing the capabilities of large language models in healthcare, we also safeguard sensitive medical data.
\section{Methods}

\subsection{Necessity of LLMs for Doctors}
Recent advancements in medical large language models (LLMs) such as PMC-LLaMA~\cite{wu2023pmc}, Med-PaLM~\cite{singhal2023large}, Med-PaLM2~\cite{singhal2025toward}, and HuatuoGPT-II~\cite{chenhuatuogpt} have significantly contributed to enhancing the domain-specific knowledge of these models and support the subsequent application of medical LLMs. Leveraging these advancements, several popular medical application models~\cite{han2023medalpaca, wang2023huatuo, bao2023disc, zhang2023huatuogpt, li2023chatdoctor, tian-etal-2024-chimed} are trained on extensive patient-doctor dialogues with the goal of functioning as autonomous virtual doctors, providing medical consultations directly to patients.

Despite advancements, the accuracy of these models in generating expert-level medical advice remains insufficient~\cite{fan2023quantifying}. Directly providing their responses to patients without medical training poses significant risks, as these patients may not be able to identify errors. For instance, a patient with suspected appendicitis presenting with abdominal pain and fever may receive an incomplete recommendation from the model, potentially delaying critical intervention.

In contrast, healthcare professionals, equipped with specialized medical knowledge, are capable of identifying such errors. This highlights the potential of developing medical large language models designed to assist doctors in addition to direct patient consultation. While recent efforts have been made to develop medical LLMs as assistants to support doctors on specific scenarios, such as MedDM~\cite{wu2023medical} for differential diagnosis and treatment recommendations and Dia-LLaMA~\cite{chen2024dia} for CT report generation. However, these works typically address only isolated tasks, leaving a significant gap in the development of LLMs capable of comprehensively supporting the full spectrum of tasks within a doctor's workflow.

\subsection{Towards Better Doctor Assistants}
Developing a medical LLM capable of assisting across the entire clinical workflow requires a dataset that comprehensively covers all relevant tasks while providing detailed and accurate responses. Furthermore, a practical benchmark is essential to evaluate whether the model can generate outputs that effectively support doctors in real-world scenarios.



\begin{table}[h]
\centering
\setlength{\tabcolsep}{3pt}
\resizebox{0.8\linewidth}{!}{%
\begin{tabular}{llcc}
\toprule
\multirow{2}{*}{\textbf{Dataset}}  &    \textbf{Applied}  &  \textbf{Entire} & \textbf{Knowledge-intensive} \\
   & \textbf{Scenarios} &  \textbf{Workflow}  & \textbf{Responses} \\
\midrule
 Huatuo-26M &  OMCD & \xmark & \cmark \\
 MedDialog  & OMCD & \xmark  & \cmark\\
 HealthCareMagic100k  & OMCD & \xmark   & \cmark\\
 ChatDoctor10k &  OMCD & \xmark & \cmark\\
 webMedQA& OMCD & \xmark &\cmark\\
 KUAKE-QIC& OMCD & \xmark &  \cmark\\
 CMeKG & KG &  \cmark  & \xmark\\
 CMExam & MCQA & \cmark & \xmark \\
 MedMCQA & MCQA & \cmark  & \xmark  \\
\hline
 DoctorFLAN & \multirow{2}{*}{DAQA} & \multirow{2}{*}{\cmark} & \multirow{2}{*}{\cmark}\\
\&DotaBench&& &\\
\bottomrule
\end{tabular}%
}
\vspace{2mm}
\caption{Comparison of existing medical training datasets. OMCD represents Online Medical Consultant Dialogue; KG represents Knowledge Graph; MCQA represents multiple-choice Question Answer; DAQA represents doctor-oriented Question Answer.}

\label{tab:dataset_comparison}
\end{table}
\textbf{Training Data Across the Entire Workflow.} 
As shown in Table \ref{tab:dataset_comparison}, existing datasets for online medical consultation dialogues, such as Huatuo-26M\cite{wang2025huatuo}, MedDialog\cite{zeng2020meddialog}, and others\cite{li2023chatdoctor, he2019applying, zhang2022cblue}, primarily provide responses for pre-diagnosis scenarios. However, these datasets only cover a limited portion of medical scenarios, making them unsuitable for comprehensive, end-to-end medical workflows. Conversely, structured resources such as knowledge graphs (e.g., CMeKG\cite{zhang2018multi}) and multiple-choice question-answer datasets (e.g., MedMCQA\cite{pal2022medmcqa} and CMExam\cite{NEURIPS2023_a48ad12d}) cover a broader range of clinical scenarios but are limited in their ability to generate knowledge-intensive, context-rich responses. Thus, there is an urgent need for a comprehensive dataset that not only encompasses the entire spectrum of a doctor's workflow but also provides detailed and context-rich answers. Such a dataset is crucial for effectively training and deploying LLMs in clinical settings.

\textbf{Doctor-Assistance Benchmark for Clinical Workflows.} 
Furthermore, existing benchmarks are insufficient for effectively evaluating models as medical assistants due to their lack of alignment with practical, real-world scenarios. Common benchmarks, such as PubMedQA~\cite{jin2019pubmedqa}, MedQA~\cite{jin2021disease}, MultiMedQA~\cite{singhal2023large}, MedMCQA~\cite{pal2022medmcqa}, CMExam~\cite{NEURIPS2023_a48ad12d}, and CMB~\cite{wang-etal-2024-cmb}, primarily focus on assessing knowledge accuracy through multiple-choice questions. However, real-world medical tasks are rarely limited to answering multiple-choice questions. Instead, they often require more nuanced decision-making accompanied by detailed analysis and explanations.  
Similarly, benchmarks like PromptCBLUE~\cite{zhu2023promptcblue}, which evaluate isolated skills such as Named Entity Recognition in medical NLP tasks, fail to capture the integrated and contextually rich requirements of doctor-assistant applications. While open-ended benchmarks like HealthSearchQA~\cite{singhal2023large} offer broader evaluations, they still fall short of covering the full spectrum of tasks encountered in a doctor’s workflow.  
Thus, there is a clear need for more realistic and comprehensive benchmarks that accurately simulate diverse medical practice scenarios. These benchmarks should be designed to evaluate the ability of LLMs to function as effective doctor assistants, providing contextually aware, detailed, and practical responses that align with real-world requirements.

\subsection{Task and Dataset Development for Clinical Workflows}
Prior clinical NLP systems such as cTAKES~\cite{savova2010mayo} have primarily focused on retrospective information extraction, aiming to standardize clinical notes through rule-based processing for tasks like concept normalization and coding. We shift the focus from retrospective extraction to prospective generation, designing workflow-aligned, open-ended tasks that reflect real-world clinical needs.
To support workflow-aligned generation, we first define a set of 22 representative tasks that span the entire clinical workflow. These tasks are derived through expert interviews and validated via a large-scale survey with licensed physicians to ensure their practical relevance and generalizability. Building on this task framework, we construct two complementary datasets: DoctorFLAN, which covers single-turn Q\&A aligned with each task, and DotaBench, which extends the task design into multi-turn dialogue settings. 


To ensure that the tasks identified align closely with the practical needs of medical professionals, we organized a symposium with 16 medical experts to discuss key tasks in the medical workflow. To avoid omissions, the experts categorize the workflow into four phases: \textbf{Pre-diagnosis}, \textbf{Diagnosis}, \textbf{Treatment}, and \textbf{Post-treatment}. In each phase, the experts identify and outline the specific tasks that doctors typically perform in daily practice. 

Pre-diagnosis tasks are actions that doctors perform before the diagnostic process. The tasks identified in this phase include \textit{Triage}, as outlined in Table \ref{tab:Diagnosis-details}. Compared to the diagnostic and treatment tasks, the pre-diagnosis tasks generally involves fewer complex medical decisions. However, the introduction of LLMs has the potential to enhance workflow efficiency by automating the generation of simple decision-making outcomes.

Diagnosis tasks encompass all activities performed by doctors during the diagnostic process that contribute to formulating the final diagnosis. The tasks are summarized in Table \ref{tab:Diagnosis-details}. Given the complexity of medical decision-making in this phase, LLMs have significant potential to assist doctors in improving decision quality. For example, in the questioning prompts task, LLMs can generate questions based on the patient's condition, encouraging doctors to conduct more comprehensive and thorough inquiries. In clinical practice, less experienced doctors may overlook critical diagnostic considerations, failing to take a complete medical history. LLMs can alleviate this by providing additional prompts that guide thorough questioning. For instance, when evaluating a patient with abdominal pain, some doctors may focus solely on the location and intensity of pain, while an LLM might prompt the doctor to inquire about changes in bowel habits, potentially revealing diagnostic clues such as irritable bowel syndrome or inflammatory bowel disease. Additionally, some tasks such as \textit{Case Summarization}, can enable LLMs to automatically generate medical case summaries, thereby saving time and effort.

Treatment tasks refer to all actions performed by doctors after diagnosis and before patient discharge. These tasks include outpatient tasks such as \textit{Medication Advice} and inpatient tasks such as \textit{Surgical Plan}, with a complete task definition provided in Table \ref{tab:Diagnosis-details}. LLMs have the potential to assist doctors in these tasks by providing advice, thereby improving decision accuracy and consistency.

Post-treatment t1q  asks are those that occur after a patient has completed their primary treatment and is transitioning to long-term recovery or ongoing management. The tasks in this phase primarily involve \textit{Health Guidance} and \textit{Follow-up Plan}, as detailed in Table \ref{tab:Diagnosis-details}. While long-term management tasks involve fewer complex decisions, they still require considerable time and effort from doctors. LLMs can help by quickly generating suggestions, improving workflow efficiency in this phase.

\begin{table*}[]
\centering
\resizebox{\textwidth}{!}{%
\begin{tabular}{l|l|l}
\toprule
\textbf{Phase} & \textbf{Specific Tasks} & \textbf{Detailed Description} \\ \midrule
Pre-diagnosis & Triage & Recommend suitable departments based on patient symptoms \\ \midrule
\multirow{11}{*}{Diagnosis} & Inquiry Prompts & Suggest follow-up questions based on patient history \\
& Symptom Inquiry & Provide key information about specific symptoms \\
& Disease Inquiry & Provide key information about specific diseases \\
& Initial Diagnosis & Identify possible conditions based on initial assessments \\
& Case Summary & Compile key points from doctor-patient dialogue into a patient case \\
& Differential Diagnosis & Differentiate between conditions with similar symptoms \\
& Next Examinations & Recommend necessary tests for further clarity \\
& Test Results Interpretation & Explain the implications of test results \\
& Definitive Diagnosis & Confirm the most likely diagnosis \\
& Disease Grading & Categorize disease severity using standard criteria \\ \midrule
\multirow{9}{*}{Treatment} & Emergency Advice & Provide guidance for urgent medical situations \\
& Treatment Plan & Propose potential treatment approaches \\
& Medication Inquiry & Offer detailed information about medications \\
& Medication Advice & Provide specific medication recommendations \\
& Complications Analysis & Highlight potential risks or complications \\
& Treatment Adjustment & Recommend updates based on patient response \\
& Surgery Necessity & Assess the need for surgical intervention \\
& Surgical Plan & Outline key considerations for surgery \\
& Preoperative Education & Explain surgery and postoperative care to patients \\ \midrule
\multirow{2}{*}{Post-Treatment} & Health Guidance & Advise on recovery and recurrence prevention \\
& Follow-up Plan & Develop a plan for regular check-ups and ongoing care\\

\bottomrule
\end{tabular}%
}
\caption{Tasks identified in the four phases.}
\label{tab:Diagnosis-details}
\end{table*}
\subsection{Validating the Task Coverage through Expert Collaboration}
\begin{figure}[t]
\includegraphics[width=\columnwidth]{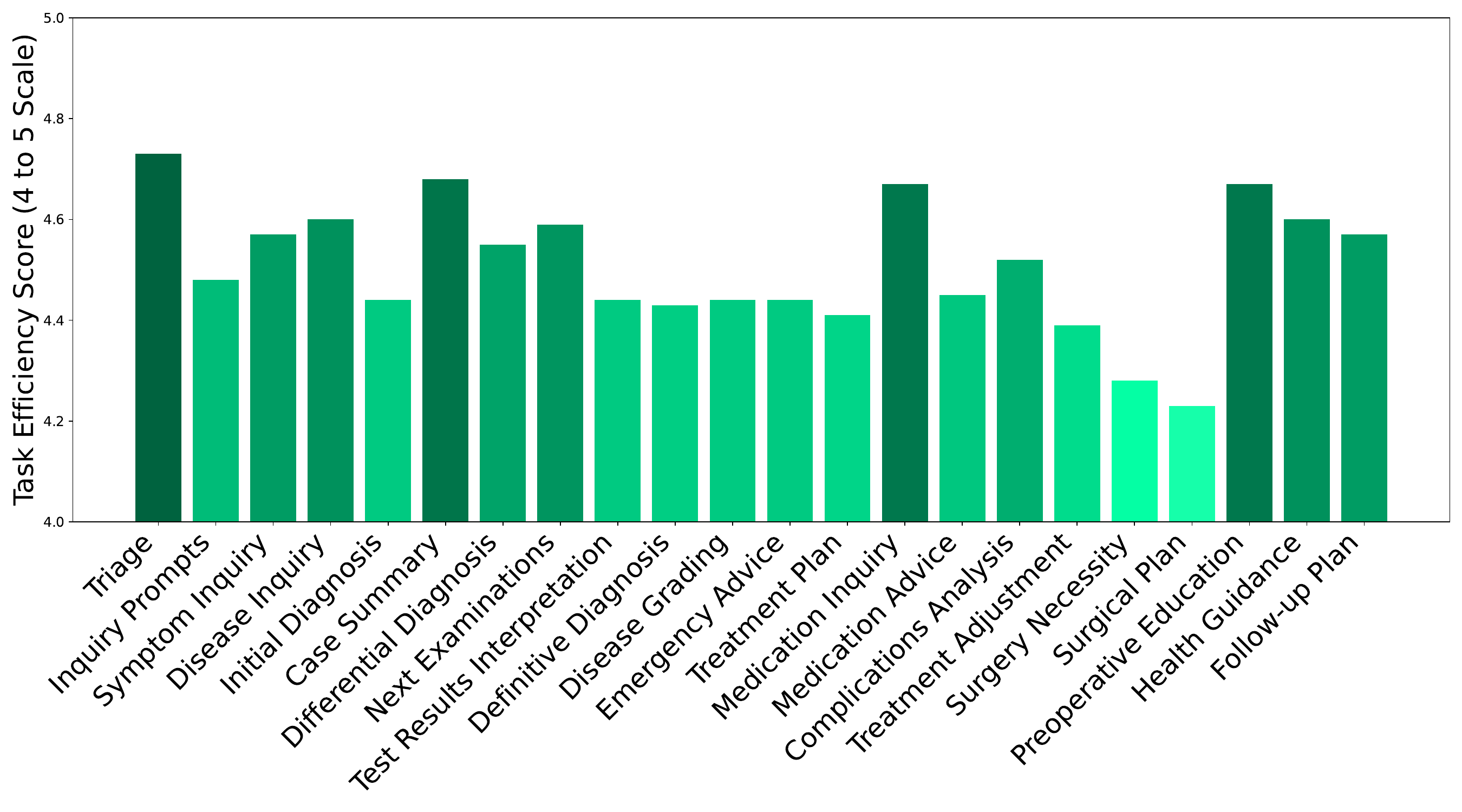}
    \refstepcounter{figure}\label{fig: importance score}
\end{figure}
To further validate the universality of the tasks defined in the focus group discussions and gain deeper insights into doctors' needs for medical LLM assistance, we conducted a survey with doctors from 13 tertiary hospitals. To ensure respondent qualifications, we distribute the survey exclusively within verified professional groups composed of licensed, practicing physicians with relevant clinical experience. The survey does not collect any personally identifiable information, in order to respect respondent privacy and encourage candid feedback. We initially list all 22 predefined tasks and ask participants to rate each task on a scale from 1 to 5, where 5 indicates that LLM assistance is crucial for improving work efficiency, and 1 signifies no impact on task efficiency. In addition, we invite the doctors to propose new tasks across four phases of their workflow, beyond the predefined tasks. Following this, we inquire about the challenges they encounter when using medical LLMs in practice, providing valuable feedback for the development of future medical assistant models.
We initially receive 82 completed questionnaires. To ensure the validity of the responses, we apply two criteria: (1) the completion time must be more than one-third of the average duration (191.82 seconds) observed across all submissions, indicating potential lack of thoughtful consideration, and (2) responses should not exhibit marked uniformity (e.g., repetitive selection of the same answer option), suggesting insufficient engagement with the content. After applying these criteria, we identify 71 valid responses for analysis.
The results reveal that most of the 22 predefined tasks receive high ratings, with scores exceeding 4, indicating that LLM assistance is highly effective for these tasks. As shown in Figure \ref{fig: importance score}, tasks such as \textit{Triage}, \textit{Case Summary}, \textit{Medication Inquiry}, and \textit{Preoperative Education} are rated particularly highly. Doctors find medical LLM assistance in these tasks especially valuable due to their repetitive nature (e.g., \textit{Case Summary, Preoperative Education}), relatively low medical risk (e.g., \textit{Triage}), and high information demands (e.g., \textit{Medication inquiry}). None of the tasks are proposed by more than five respondents, reinforcing that the final set of 22 tasks is widely applicable and relevant across the surveyed doctors.
Among the participants, 46.5\% report using LLMs to assist with their clinical work. When asked about the limitations of current medical LLM capabilities, respondents show strong consensus on several issues. Specifically, 42.2\% of doctors identify problems with noncompliance to instructions, 48.5\% report instances of incorrect answers, and 39.4\% express concerns about the LLM’s inability to provide accurate references. Additionally, doctors emphasize the necessity of continuously updating the LLM's knowledge base and incorporating self-correction mechanisms to improve the reliability and accuracy of the model’s outputs.

\subsection{Task Comparison Between Typical Medical Datasets and Our Defined Tasks}
\begin{figure}
    \centering
    \includegraphics[width=0.8\columnwidth ]{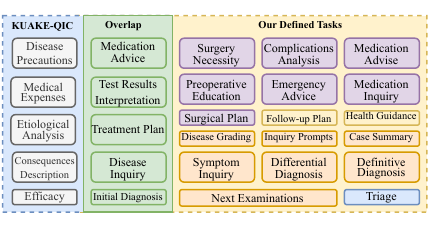}
    \refstepcounter{figure}\label{fig:task_overlap}
\end{figure}
We further compare the tasks defined in our framework with those in typical medical datasets, using KUAKE-QIC~\cite{zhang2022cblue} as a representative example. While some overlap exists between the datasets, our defined tasks introduce 17 additional tasks not covered by KUAKE-QIC, highlighting the broader scope and versatility of our approach, as illustrated in Figure \ref{fig:task_overlap}.

\subsection{DoctorFLAN Construction}
\begin{figure*}[t]
\includegraphics[width=\textwidth]{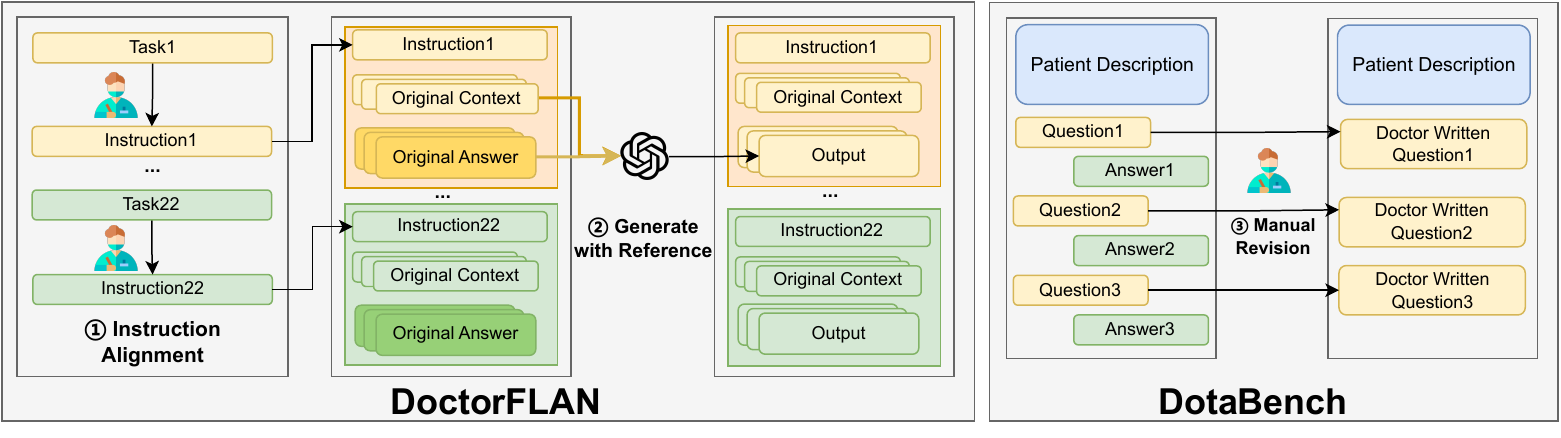}
    \refstepcounter{figure}\phantomcaption\label{fig: data_process}
\end{figure*}
To create a comprehensive dataset covering the entire clinical workflow, we construct single-turn DoctorFLAN based on the 22 predefined tasks. First, we collect raw medical data from a variety of sources, then we heuristically filter and map the data to the relevant tasks. The dataset is refined in two stages: instruction normalization and response polishing. Following the initial construction, we conduct manual verification of a subset of the data by medical experts to ensure its quality, as shown in Figure \ref{fig: data_process}.

\textbf{Data Source.} 
We use three primary data sources: medical multiple-choice questions (MCQs) (e.g., https://www.medtiku.com/), medical encyclopedia entries (e.g., https://m.120ask.com/), and high-quality existing medical datasets such as PromptCBLUE~\cite{zhu2023promptcblue}. MCQs are chosen for their ability to simulate a broad range of clinical scenarios, making them highly relevant to real-world practice. The medical encyclopedia, which contains detailed information on topics such as drugs and symptoms, provides a comprehensive and reliable reference, especially for tasks like \textit{Medication Inquiry}. Additionally, we include overlapping datasets from resources like the \textit{Case Summary} subset in PromptCBLUE.

\textbf{Preprocessing and Task Mapping.}
After collecting the raw data, we perform deduplication using Jaccard similarity (threshold = 0.8) to eliminate near-duplicate entries and improve data quality~\cite{lee2022deduplicating}. We then categorize the data into the 22 predefined task types using a carefully designed set of task-specific regular expressions. Each task is associated with multiple regex patterns, which are iteratively refined based on expert feedback. In each iteration, we sample 50 examples for manual annotation by a senior physician to assess classification quality. The refinement process continues until the regex-based categorization achieves over 95\% agreement with expert labels, ensuring high precision and consistency. The description of the regex process and an example for task classification are provided in Section A of the Supplementary Information.

\begin{table*}[h!]
\centering
\resizebox{\linewidth}{!}{
{\scriptsize 
\begin{tabular}{l}
\toprule
\textbf{Instruction:}\\
Facial pain: Provide a detailed overview of the symptoms, possible causes, and related \\ diseases. Include the definition of the symptoms, typical manifestations, common causes \\that may cause the symptoms, and other health issues related to these symptoms.\\\midrule
\textbf{Reference:}\\
associated\_diseases: Infection in the infratemporal space, chronic maxillary sinusitis, \\temporal arteritis, infection under the lower jaw, infection in the submandibular space, \\osteomyelitis of the jaw, bone cyst, petrositis of the temporal bone, acute osteomyelitis \\
of the jaw, acute submandibular sialadenitis\\
overview: Facial pain refers to pain on one or both sides of the face, or at the forehead. \\
This pain can be dull or throbbing and can be extremely sharp.Facial pain or spasms, \\
generally not life-threatening, but the sufferer is very miserable; even if the pain\\ is bearable, the facial expression appears unnatural.\\ 
cause: Facial pain is usually caused by local tissue infection or inflammation; some are \\throbbing pain, some are severe stabbing pain, and the presented symptoms vary. Muscle \\ inflammation is the most familiar cause of facial pain, also known as temporomandibular \\joint disorder. Followed by trigeminal neuralgia, herpes zoster. It can also be seen in facial \\pain caused by oral inflammation, and pain caused by local trauma.\\\midrule
\textbf{Answer:}\\
Facial pain is a relatively common clinical symptom, characterized by various forms and \\manifestations. Patients may describe facial pain as dull pain, throbbing pain, burning \\sensation, or extreme stabbing pain. When considering the causes of facial pain, it is \\necessary to comprehensively consider the distribution, nature, triggering and alleviating \\factors, and accompanying symptoms.\\
According to \textless textbook knowledge\textgreater, diseases associated with facial pain \\include, but are not limited to, infection in the infratemporal space, chronic maxillary \\sinusitis, temporal arteritis, infection under the lower jaw, infection in the submandibular \\space, and osteomyelitis of the jaw.
\\\midrule
\textbf{Doctor Comment:}\\
The answer lacks specific detail compared to the original reference answer.\\
\bottomrule
\end{tabular}
}
}
\vspace{2mm}
\caption{The content presented has been translated from Chinese: An example of a response judged by doctors to be less practical compared to the original, illustrating a potential challenge in ensuring the utility of LLM-generated content in medical data.}
\label{tab:verification failure}
\end{table*}
\textbf{Reference-enhanced Refinement.}
Although we have gathered data for the 22 tasks, the initial dataset contains issues such as poorly worded instructions and overly brief responses. To address these problems, we implement a two-step refinement process: instruction alignment and response polishing. In the instruction alignment phase, medical professionals are enlisted to manually draft task-specific instructions for each of the 22 tasks, ensuring that the instructions accurately reflect real-world clinical scenarios and align with the intended task. In the response polishing phase, we ask GPT-4 to generate more comprehensive responses by referencing the original data to enhance their quality. The final dataset contains 92350 samples, divided into a training set DoctorFLAN-\textit{train} and a test set DoctorFLAN-\textit{test}. The test set includes 25 randomly sampled entries from each task, for a total of 550 samples.

To ensure that the responses generated by GPT-4 are factually accurate and realistic, we use a structured review process in which a sample of 1050 responses (50 samples per item across 22 tasks) are reviewed by three medical professionals, each reviewing 350 items. The verification process is overseen by a senior expert with a high-level title, who has dedicated 10 hours to ensure a thorough assessment. Each model response is reviewed alongside its corresponding reference answer, and the reviewers are instructed to revise or refine the outputs as needed based on that reference. Rather than conducting blind, independent annotation, this process is designed as a reference-grounded refinement task aimed at improving factual correctness and clinical appropriateness. This approach balances thoroughness with practical limitations, ensuring credible verification within the available resources.
The verification criteria include \textit{Correctness}, where a response is considered correct if it contains no factual errors, and \textit{Practicality}, where a response is deemed practical if it is more effective than the original answer. Our results demonstrate correctness (100\%) and practicality (99.9\%), underscoring the robustness of the DoctorFLAN. 
In a detailed examination of the data verification stage, we identify an instance where a doctor noted the lack of practicality, commenting on the "lack of specific details," as shown in Table \ref{tab:verification failure}. Such feedback suggests that the responses refined by GPT-4 can sometimes fall short in complex practical medical contexts, highlighting an area for future improvement. 

\subsection{DotaBench Construction}
Extending the single-turn dataset DoctorFLAN, we introduce multi-turn DotaBench to evaluate multi-turn dialogues involving medical assistants. This extension is motivated by the need to assess an LLM's ability to operate in realistic clinical settings, where conversations often span multiple turns and involve a sequence of logically connected questions. While DoctorFLAN captures isolated queries, DotaBench focuses on multi-turn interactions in which each question is designed to build upon the previous one, simulating the stepwise inquiry process commonly used by physicians in real-world consultations.

\textbf{Data Source.} 
To ensure clinical authenticity, we select CMB-Clin~\cite{wang-etal-2024-cmb} as the source corpus. CMB-Clin is a multi-round question-answering dataset derived from real medical records. However, its original format consists of 2–4 standalone Q\&A pairs that lack contextual continuity, making it unsuitable for dialogue-based evaluation in its raw form.

\textbf{Reference-enhanced Refinement.}
To address this limitation, we work with licensed physicians to manually restructure the data into coherent three-turn dialogues. Specifically, we extract key clinical elements from each case, such as chief complaints, physical findings, and diagnostic test results, and ask physicians to reformulate them into contextually connected questions that reflect realistic consultation workflows. The original answers from CMB-Clin are retained as reference responses, which are later used to support reference-based evaluation under the LLM-as-a-judge framework. A representative example illustrating this transformation is included in Supplementary Tables 1 and 2. Unlike DoctorFLAN, which directly involves LLMs in data generation, DotaBench is crafted without LLM intervention, thereby eliminating the need for subsequent data verification and ensuring controlled evaluation conditions.

\begin{table}
\centering
\small
\begin{tabular}{@{}lccc@{}}
\toprule
    & \multicolumn{2}{c}{\textbf{DoctorFLAN}} & \textbf{DotaBench} \\
\midrule
\textbf{Type}          & \multicolumn{2}{c}{Single-turn}      & 3-turns           \\
\textbf{Split}         & train              & test               & test              \\
\textbf{Specialist}    & 27 &27              &  -          \\
\textbf{Task}          & 22    &22         &   -       \\
\textbf{\#Q/task}  & -*                  & 25                 & -                 \\
\textbf{\#Q in total}  & 91,776             & 550                & 74                \\
\bottomrule
\end{tabular}
\vspace{2mm}
\caption{The Statistics of DoctorFLAN and DotaBench Dataset.}
\label{tab:dataset-summary}
\end{table}
\begin{table*}[]
\centering
\resizebox{\textwidth}{!}{%
\begin{tabular}{l|llll}
\toprule
\multirow{4}{*}{Specialist}  
& Gastroenterology, Pediatrics, Obstetrics \& Gynecology, Respiratory,  \\&Medicine Cardiology, Neurology, General Surgery Stomatology, Nephrology, \\& Hepatology, Orthopedics,  Urology, Spine Surgery, Cardiothoracic Surgery, \\& OphthalmologyHematology, Endocrinology, Oncology, Emergency Medicine, \\ 
& Infectious Disease, Traditional Chinese MedicineRheumatology \& Immunology, \\& Neurosurgery, Dermatology, Otorhinolaryngology (ENT), Vascular Surgery, \\& Multidisciplinary \\ \midrule
\multirow{6}{*}{Task}     &  \textbf{Pre-Diagnosis:} Triage \\
& \textbf{Diagnosis:} Inquiry Prompts, Symptom Inquiry, Disease Inquiry, Initial Diagnosis, \\ & 
Case Summary Differential Diagnosis, Next Examinations, Test Results Interpretation, \\ &Definitive Diagnosis, Disease Grading   \\ & 
\textbf{Treatment:} Emergency Advice, Treatment Plan, Medication Inquiry, \\ & Medication Advice, Complications Analysis  Treatment Adjustment, \\ &Surgery Necessity,  Surgical Plan, Preoperative Education \\ &
\textbf{Post-Treatment:} Health Guidance, Follow-up Plan  \\ \bottomrule
\end{tabular}%
}
\caption{Specialists and Tasks in the DoctorFLAN Dataset.}
\label{tab:dataset-details}
\end{table*}
\subsection{Data Statistic}

The statistical analysis of the DoctorFLAN and DotaBench datasets is presented in Table \ref{tab:dataset-summary}.
The DoctorFLAN dataset comprises 92,326 instances across 22 distinct tasks, involving 27 medical specialties in total as detailed in Table \ref{tab:dataset-details}, demonstrating the comprehensive coverage of DoctorFLAN in real clinical scenarios. 
In addition, we have extracted a subset of 25 instances from each task, referred to as DoctorFLAN-\textit{test} for evaluation. The training and test sets are created via random split. The DotaBench dataset includes 74 instances of 3-turn conversations.

\subsection{Model Training}
We fine-tune two open-source backbone models, Yi-6B and Baichuan2-7B-Base, using a standard supervised fine-tuning (SFT) framework with an autoregressive, decoder-only architecture. To ensure the model captures both domain-specific expertise and general ability, we construct a mixed training corpus comprising 92k task-aligned medical samples from DoctorFLAN, 101k general-purpose instruction samples from datasets such as Evol-instruct{~\cite{xu2023wizardlm}}, ShareGPT~\cite{sharegpt}, and 51k additional medical QA pairs from CMExam~\cite{NEURIPS2023_a48ad12d}. 

All models are trained on 4 NVIDIA A100 GPUs. We set the maximum input sequence length to 4096 tokens and used a per-GPU batch size of 4, training for 3 epochs with a learning rate of $5 \times 10^{-5}$. The optimization used the AdamW optimizer with decoupled weight decay, and gradient checkpointing is enabled to reduce memory consumption. Mixed precision training is performed using fp16 format to accelerate computation.

The objective function is the negative log-likelihood (NLL) of the target response given the prompt, encouraging the model to generate accurate and fluent outputs aligned with medical task instructions. Specifically, the loss is defined as:

\begin{equation}
\mathcal{L}_{\mathrm{SFT}} = - \sum_{t=1}^{T} \log P(y_t \mid x, y_{<t})
\label{eq:sft_loss}
\end{equation}

where $x$ denotes the input prompt and $y_t$ the target token at time step $t$. The final model checkpoint was selected after three training epochs based on manual review and preliminary validation performance, without using early stopping or automated selection heuristics.



\subsection{Evaluation Models}
To comprehensively evaluate the performance of medical-specific models trained on various backbones and datasets, we assess a wide range of Chinese medical LLMs on DoctorFLAN-\textit{test} and DotaBench. 

Among the domain-specific models, we include BianQue-2~\cite{chen2023bianque}, a medical model fine-tuned from ChatGLM-6B~\cite{zengglm} using patient-doctor dialogues; DISC-MedLLM~\cite{bao2023disc}, a model based on the Baichuan-13B-Base architecture designed for deep medical interactions; HuatuoGPT-7B~\cite{zhang2023huatuogpt}, fine-tuned from Baichuan-7B for Chinese medical consultation; and HuatuoGPT-II-7B~\cite{chenhuatuogpt}, a state-of-the-art medical LLM built on Baichuan2-7B with extensive medical knowledge.

We also evaluate general-purpose models to provide a performance baseline. These include Qwen-1.8B-Chat~\cite{bai2023qwen}, fine-tuned with supervised fine-tuning (SFT) and reinforcement learning with human feedback (RLHF); Baichuan-13B-Chat~\cite{baichuan13b}, which shares the same backbone as DISC-MedLLM and demonstrates strong general performance; and Baichuan2 models including Baichuan2-7B-Chat and Baichuan2-13B-Chat~\cite{yang2023baichuan}. We further include Yi-6B-Chat and Yi-34B-Chat~\cite{young2024yi}, which represent two scales of models from the Yi series, comparable to Qwen and Baichuan.

To broaden the comparison, we additionally report results from proprietary models such as GPT-3.5, GPT-4, and Claude-3.

All models are evaluated using the same decoding hyperparameters: \texttt{max\_new\_tokens = 1024}, \texttt{top\_p = 0.7}, \texttt{temperature = 0.5}, and \texttt{repetition\_penalty = 1.1}. We adopt Chain-of-Thought prompting, without using any additional augmentation techniques.

\subsection{Evaluation Method}

Considering both accuracy, reliability, and cost, our evaluation methodology incorporates both automatic and human evaluations.

\textbf{Automatic Evaluation.}
Our task involves open-ended answer generation in medical contexts, where multiple correct and clinically valid responses may exist. In such settings, traditional metrics such as BLEU and ROUGE, which rely on N-gram overlap with reference answers, are often inadequate. These metrics fail to capture semantic consistency when answers are phrased differently yet medically equivalent, and are also highly sensitive to variations in response length. To address these limitations, we employ GPT-4 (gpt-4-0125-preview) for automatic evaluation, a method shown to be highly effective in previous research ~\cite{liu2023g}. To ensure evaluation accuracy, we adopt a reference-based model evaluation approach, where the LLM refers to the provided reference and scores responses based on predefined criteria. These scoring standards include: Accuracy (assessing the correctness and reliability of the information), Coherence (evaluating the clarity and logical flow of the responses), Relevance (measuring how closely each response addresses the prompt), and Thoroughness (judging the depth and completeness of the response in covering the topic). During evaluation, we apply Chain-of-Thought (CoT) prompting both in response generation and in the LLM-as-a-judge scoring process. We do not use any external augmentation techniques, such as retrieved rationales or tool-assisted reasoning. The evaluation is performed using GPT-4, accessed via the official OpenAI API with default inference settings.
To support reproducibility, we provide the full evaluation prompt in Supplementary Figures 1 and 2.

To balance accuracy and resource constraints, we conduct human evaluation on a subset of models. For DoctorFLAN-\textit{test}, which contains 550 questions in total, we divide them into six roughly equal parts, with 91 or 92 questions per evaluator. Each evaluator is assigned a set of questions and tasked with rating the responses of all six models for each question, ensuring a fair and consistent evaluation across all models. The evaluation team consists of six healthcare professionals with varying levels of experience: three mid-level professionals with 5-6 years of experience, two associate senior professionals with 12 years of experience, and one senior professional with 26 years of experience. Evaluators are compensated based on their professional seniority, with senior professionals receiving an hourly rate of 250 RMB, while mid-level professionals are paid 165 RMB per hour.
For DotaBench, we invite three doctors to participate in the evaluation process, with each spending an average of 3 hours reviewing the data.

\section*{Data Availability}
DoctorFLAN and DotaBench datasets used in this study are available at https://huggingface.co/datasets/FreedomIntelligence/DoctorFLAN and https://huggingface.co/datasets/FreedomIntelligence/DotaBench, respectively.

\section*{Code Availability}
The source code for DotaGPT training and evaluation is available at https://github.com/FreedomIntelligence/DotaGPT.

\section*{Acknowledgements}
This work was supported by the Major Frontier Exploration Program (Grant No. C10120250085) from the Shenzhen Medical Academy of Research and Translation (SMART), the Shenzhen Science and Technology Program (JCYJ20220818103001002), Shenzhen Doctoral Startup Funding (RCBS20221008093330065), Tianyuan Fund for Mathematics of National Natural Science Foundation of China (NSFC) (12326608), Shenzhen Science and Technology Program (Shenzhen Key Laboratory Grant No. ZDSYS20230626091302006), and Shenzhen Stability Science Program 2023, Shenzhen Key Lab of Multi-Modal Cognitive Computing. 

\section*{Author Contributions}
W.X. led the task survey design, constructed the dataset, conducted the main experiments, and drafted the initial manuscript.
J.F. and B.W. proposed the original idea and, together with A.G., P.T., and X.W. (Xiang Wan), made substantial contributions to manuscript revision.
Q.X. led the expert survey and coordinated the human evaluation.
Y.Z. contributed to early-stage data construction.
X.W. (Xidong Wang) developed the DotaBench dataset.
J.C. helped refine the data construction methodology, and K.J. supported data analysis.
All authors reviewed and approved the final manuscript.

\section*{Competing Interests}
The authors declare no competing interests.
\section*{Reporting Checklist}
This study does not involve clinical trials or systematic reviews. Therefore, reporting checklists such as PRISMA or CONSORT are not applicable. 
\bibliography{sn-bibliography}
\section*{Figure Legends}
\textbf{Figure 1.} Task Categories Finalized for LLMs in Medical Assistance, Organized by Four Phases: Pre-Diagnosis, Diagnosis, Treatment, and Post-Treatment.\\
\textbf{Figure 2.2.} Correlations between human and automatic evaluations on DoctorFLAN-test, illustrating task-level consistency.\\
\textbf{Figure 4.4.} Comparative assessment of task efficiency scores for each task according to our survey.\\
\textbf{Figure 4.5.} Task overlap between our defined tasks and KUAKE-QIC, highlighting the 17 unique tasks introduced in our framework.\\
\textbf{Figure 4.6.} Reference-Enhanced Refinement in DoctorFLAN and DotaBench.
\appendix
\setcounter{figure}{0}
\setcounter{table}{0}
\renewcommand{\thefigure}{\arabic{figure}}
\renewcommand{\thetable}{\arabic{table}}
\captionsetup[figure]{name=Supplementary Figure}
\captionsetup[table]{name=Supplementary Table}

\section{Regex-Based Data Categorization}
\label{appendix:regex}
\textit{Note: Since our QA corpus is in Chinese, the regex patterns are expressed using Chinese characters where necessary. Each pattern is accompanied by an English explanation for clarity.}

To enable efficient and scalable preprocessing, we employ regular expression (regex) rules to automatically filter and categorize medical QA samples. The overall pipeline consists of two stages: (1) general data cleaning and (2) task-specific classification using field-wise regex matching.

\paragraph{General Filtering.}
Before task categorization, we applied two regex-based filters to the entire dataset:
\begin{itemize}
    \item \textbf{Case-based QA Selection:} We retained only questions that involve real-world case descriptions, identified by matching sentence openings such as “患者”, “男”, “女”, “患儿”, or “某患者”.
    \item \textbf{Exclusion of Image/Table-Based Questions:} We removed samples referencing visual content using the following regex pattern:
    \begin{tcolorbox}[colback=lightgray!10,colframe=black, title={Regex Rule for Filtering Text-Based Content}]
    \texttt{r'如图|结合图像|图像如下|img|图示|图[0-9]+|表[0-9]+'}
    \end{tcolorbox}
\end{itemize}
\paragraph{Example: Category Classification via Multi-Field Regex Matching.}
Each QA item was decomposed into five semantic fields to support modular rule design:
\begin{enumerate}
    \item Case description
    \item Question content
    \item Metadata (e.g., subject area such as “传染病学” / infectious diseases)
    \item Answer text
    \item Answer Option Set
\end{enumerate}

To classify data into specific clinical categories, we applied regex rules over \textbf{multiple fields in combination}, instead of relying on any single field alone.

For instance, to extract QA samples related to \textit{Differential Diagnosis} task, we used the following logic:
\begin{itemize}
    \item \textbf{Field 2 (Question Content):} We matched the question text against a regex that detects diagnostic comparison terms (e.g., “鉴别”, “诊断”, “区别”), while filtering out questions focused on treatment, symptoms, or procedures. The regex used was:
    \begin{tcolorbox}[colback=lightgray!10, colframe=black, title={Regex on Question Content}]
    \verb+r'^(?!.*(?:明确|检查|并发症|意见|利于|必备条件|鉴别要点|治疗|问题|项目|体征|条件))(?=.*(?:鉴别|诊断|区别)).*$'+
    \end{tcolorbox}

    \item \textbf{Field 5 (Answer Option Set):} We required that the question includes more than one answer option, as differential diagnosis questions often present multiple candidate conditions for selection.
\end{itemize}
Only when both conditions were satisfied, namely that relevant keywords appeared in the question and that multiple answer options were present, did we label the sample as belonging to the \textit{Differential Diagnosis} category.

This combination-based rule design ensures higher precision and flexibility, and can be extended to other categories by customizing field-specific regex patterns.

\section{Dotabench Construction Details}
\label{appendix:dota_example}


\begin{table*}[]
\centering
\resizebox{\textwidth}{!}{%
\begin{tabular}{l|l}
\toprule
\textbf{Prompt} & \textbf{Question}  \\ \midrule
\multirow{5}{*}{Turn 1} 
& 这位49岁的男性病人在3小时前解大便后出现右下腹疼痛，他自己可以触及右下腹的一个包块。\\&他之前都很健康，没有什么特别的既往史。你能帮我根据这些信息给出一个初步的诊断吗？ \\
& (Translated) This 49-year-old male patient developed right lower abdominal pain three hours\\& ago after a bowel movement, and he can palpate a lump in the same area. He has no notable\\&  past medical history.Based on this information, could you provide a preliminary diagnosis?
\\ \midrule
\multirow{7}{*}{Turn 2} & 我刚刚进行了体格检查，发现他体温37.8℃，心率101次／分，呼吸22次/分，血压100/60mmHg。\\
&于右侧腹股沟区可扪及一圆形肿块约4cm×4cm大小，并有压痛、界欠清，在腹股沟韧带上内方。\\
&请问还需要哪些辅助检查来确定诊断？\\
&(Translated) I just performed a physical examination. The patient has a temperature of 37.8°C, \\
&a heart rate of 101 bpm, respiratory rate of 22 breaths/min, and blood pressure of 100/60 mmHg. \\
&A round mass approximately 4 cm × 4 cm is palpable in the right inguinal region, with tenderness\\
& and poorly defined borders. It is located medial to the inguinal ligament. What additional diagnostic\\
&tests would you recommend to confirm the diagnosis?\\ \midrule
\multirow{10}{*}{Turn 3} &
检验结果出来了。血常规显示白细胞计数5.0×109／L，中性粒细胞78％。尿常规正常。多普勒\\
&超声检查沿腹股沟纵切可见一多层分布的混合回声区，宽窄不等，远端膨大，边界整齐，长约\\
&4～5cm。腹部X线检查可见阶梯状液气平。根据这些信息，请帮我明确诊断，并提供治疗方案。\\
&(Translated) The test results are now available. The complete blood count shows a white blood\\
&cell count of 5.0×10⁹/L with 78\% neutrophils. Urinalysis is normal. Doppler ultrasound reveals\\
&a multilayered mixed-echo region along the longitudinal section of the inguinal area, with \\
&uneven width and distal enlargement, measuring approximately 4–5 cm with well-defined borders.\\
&Abdominal X-ray shows a step-ladder pattern of air-fluid levels. Based on this information, \\
&could you confirm the diagnosis and recommend a treatment plan?\\\bottomrule
\end{tabular}%
}
\caption{A manually constructed DotaBench example consisting of three contextually linked turns that reflect realistic consultation
workflows.}
\label{tab:dotabench_example}
\end{table*}

\begin{table*}[]
\centering
\resizebox{\textwidth}{!}{%
\begin{tabular}{l|l}
\toprule
\multirow{18}{*}{Case Description} 
& 现病史（1）病史摘要 病人，男，49岁，3小时前解大便后出现右下腹疼痛，右下腹可触及\\
&一包块，既往体健。（2）主诉 右下腹痛并自扪及包块3小时。 体格检查 体温： T 37.8℃，\\
&P 101次／分，呼吸22次/分，BP 100/60mmHg，腹软，未见胃肠型蠕动波，肝脾肋下未及，\\
&于右侧腹股沟区可扪及一圆形肿块，约4cm×4cm大小，有压痛、界欠清，且肿块位于腹股沟\\
&韧带上内方。辅助检查（1）实验室检查 血常规：WBC 5.0×109／L，N 78％。 尿常规\\
&正常。 （2）多普勒超声检查 沿腹股沟纵切可见一多层分布的混合回声区，宽窄不等，远端\\
&膨大，边界整齐，长约4～5cm。（3）腹部X线检查可见阶梯状液气平。\\
&(Translated) Present Illness History: (1) Summary: A 49-year-old male developed right lower\\ &abdominal pain three hours ago after defecation, with a palpable mass in the same area. \\
&No significant medical history. (2) Chief Complaint: Right lower abdominal pain and a \\
&self-palpated mass for 3 hours. Physical Examination: Temperature: 37.8°C, Pulse: 101 bpm,\\
&Respiration: 22/min, Blood Pressure: 100/60 mmHg. Abdomen soft, no visible peristaltic \\ &waves, liver and spleen not palpable. A round mass (~4×4 cm) with tenderness and poorly \\
&defined borders is palpable in the right inguinal region, medial to the inguinal ligament. \\ &Auxiliary Tests: (1) Laboratory Tests: CBC: WBC 5.0×10⁹/L, Neutrophils 78\% Urinalysis: \\ &Normal (2) Doppler Ultrasound: Multilayered mixed-echo area along the inguinal longitudinal \\
&section with variable width and distal enlargement; well-defined borders; ~4–5 cm in length \\
&(3) Abdominal X-ray: Step-ladder air-fluid levels observed
\\ \midrule
\multirow{2}{*}{Question 1} & 简述该病人的诊断及诊断依据。\\
& (Translated) Summarize the diagnosis and diagnostic rationale.\\\midrule
\multirow{2}{*}{Question 2} &简述该病人的鉴别诊断。\\
&(Translated) Summarize the differential diagnosis.\\\midrule
\multirow{2}{*}{Question 3} &简述该病人的治疗原则。\\
&(Translated) Summarize the treatment principles.\\\bottomrule
\end{tabular}%
}
\caption{The original CMB-Clin case record used as the source for DotaBench construction. Note that the original QA pairs are isolated and lack multi-turn context.}
\label{tab:cmb_example}
\end{table*}

Supplementary Tables \ref{tab:dotabench_example} and  \ref{tab:cmb_example} present an example of how we convert a raw case from CMB-Clin into a contextually linked multi-turn consultant sample in DotaBench.

\section{Evaluation Prompt}
\label{appendix:evaluation_prompt}
\begin{figure}[t]
\begin{AIbox}[colback=lightgray!10]{Evaluation Prompt for DoctorFLAN-\textit{test}}
{\bf System Prompt:} \\
{\footnotesize
Please act as an impartial judge and evaluate the quality of the response provided by an AI assistant to the user question displayed below. \\
Requirements: Your assessment should focus primarily on the consistency between the assistant's answer and the reference answer. \\
Begin your evaluation by providing a short explanation. Be as objective as possible. After providing your explanation, you must rate the response on a scale of 1 to 10 by strictly following this format: "\lbrack \lbrack Rating \rbrack \rbrack", for example: "Rating: \lbrack \lbrack 5\rbrack \rbrack".\\
}
{\bf Prompt:} \\
{\footnotesize
\lbrack Question \rbrack \\
\{question\}\\
\\
\lbrack The Start of Reference Answer\rbrack \\
\{reference\}\\
\lbrack The End of Reference Answer\rbrack \\
\lbrack The Start of Assistant's Answer]\\
\{answer\}\\
\lbrack The End of Assistant's Answer \rbrack
}
\end{AIbox}
\caption{Evaluation Prompt for DoctorFLAN-\textit{test}.}
\label{fig:prompt_doctorflan}
\end{figure}
\begin{figure}[t]
\begin{AIbox}[colback=lightgray!10]{Evaluation Prompt for DotaBench}
{\bf System Prompt:} \\
{\footnotesize
Please act as an impartial judge and evaluate the quality of the response provided by an AI assistant to the user question displayed below. \\
Requirements: Your assessment should focus on the overall quality of the responses based on the following criteria:\\
Accuracy: Evaluate the correctness and reliability of the information provided.\
Coherence: Assess the clarity and logical flow of the responses.\
Relevance: Determine how closely each response addresses the question asked.\
Thoroughness: Judge the depth and completeness of the response in covering the topic. \\
You will be given the assistant's answer and some references. The reference consists of Q\&A pairs related to the patient, which are completely accurate and can be used as a reliable source of truth.\\
You evaluation should focus on the assistant's answer to the first question. Begin your evaluation by providing a short explanation. Be as objective as possible. After providing your explanation, you must rate the response on a scale of 1 to 10 by strictly following this format: "\lbrack \lbrack Rating \rbrack \rbrack", for example: "Rating: \lbrack \lbrack 5\rbrack \rbrack".\\
}
{\bf Prompt:} \\
{\footnotesize
<|The Start of Reference|>\\
\\
\{reference\}\\
\\
<|The End of Reference|>\\
\\
\\
<|The Start of Assistant A's Conversation with User|>\\
\\
\#\#\# User:\\
\{question\_1\}\\
\\
\\
\#\#\# Assistant A:\\
\{answer\_1\}\\
\\
<|The End of Assistant A's Conversation with User|>}
\end{AIbox}
\caption{Evaluation Prompt for DotaBench}
\label{fig:prompt_dotabench}
\end{figure}
Supplementary Figures \ref{fig:prompt_doctorflan} and \ref{fig:prompt_dotabench} show the complete prompt templates used for DoctorFLAN and DotaBench, respectively.

\end{CJK}
\end{document}